\documentclass[runningheads]{llncs}
\usepackage{stfloats}
\usepackage{bm}
\usepackage{amsmath}
\usepackage{amsfonts}
\usepackage{booktabs}
\usepackage{array}
\usepackage{cite}
\usepackage{graphicx}
\usepackage{enumerate}
\usepackage{subfigure}
\usepackage{bbding}
\usepackage[colorlinks,
            linkcolor=blue,
            anchorcolor=blue,
            citecolor=blue]{hyperref}
\begin{document}
%Image Captioning with Text-dependent Attention and External knowledge
\title{Boost Image Captioning with Knowledge Reasoning
% \thanks{This work is supported by the National Natural Science Foundation of
% China (Nos. 61966004, 61663004, 61866004, 61762078), the Guangxi Natural 
% Science Foundation (Nos. 2019GXNSFDA245018, 2018GXNSFDA281009, 2017GXNSFAA198365),
% the Guangxi “Bagui Scholar” Teams for Innovation and Research Project, the
% Guangxi Talent Highland Project of Big Data Intelligence and Application,
% Guangxi Collaborative Innovation Center of Multi-source Information 
% Integration and Intelligent Processing.}
}
% begin
\author{Feicheng Huang\inst{1} \and
Zhixin Li\inst{1, (}\Envelope\inst{)} \and
Haiyang Wei\inst{1} \and Canlong Zhang\inst{1} \and 
Huifang Ma\inst{2}}
\authorrunning{Huang et al.}
\institute{Guangxi Key Lab of Multi-source Information Mining and Security,\\ 
Guangxi Normal University, Guilin 541004, China \\
\and
College of Computer Science and Engineering, Northwest Normal University, \\
Lanzhou 730070, China\\
\email{lizx@gxnu.edu.cn
}}
% end
% \author{\IEEEauthorblockN{Feicheng Huang$^1$, Zhixin Li$^{1,*}$, Shengjia Chen$^1$, Canlong Zhang$^1$, Huifang Ma$^{1,2}$}
% 	\IEEEauthorblockA{$^1$Guangxi Key Lab of Multi-source Information Mining and Security,\\ Guangxi Normal University, Guilin 541004, China\\
% 		$^2$College of Computer Science and Engineering,  Northwest Normal University, Lanzhou 730070, China\\
% 		$^*$Corresponding Author. E-mail: lizx@gxnu.edu.cn}}

\maketitle              % typeset the header of the contribution
\begin{abstract}
  Automatically generating a human-like description for a given image is a
  potential research in artificial intelligence, which has attracted a great
  of attention recently. Most of the existing attention methods explore the
  mapping relationships between words in sentence and regions in image, such
  unpredictable matching manner sometimes causes inharmonious alignments that
  may reduce the quality of generated captions. In this paper, we make our
  efforts to reason about more accurate and meaningful captions. We first propose
  word attention to improve the correctness of visual attention when generating
  sequential descriptions word-by-word. The special word attention emphasizes
  on word importance when focusing on different regions of the input image,
  and makes full use of the internal annotation knowledge to assist the calculation
  of visual
  attention. Then, in order to reveal those incomprehensible intentions
  that cannot be expressed straightforwardly by machines, we introduce a new strategy to
  inject external knowledge
  extracted from knowledge graph into the encoder-decoder framework to
  facilitate meaningful captioning. Finally, we validate our model on two freely
  available captioning benchmarks: Microsoft COCO dataset and Flickr30k dataset. The
  results demonstrate that our approach achieves state-of-the-art performance
  and outperforms many of the existing approaches.

\end{abstract}
\section{Introduction}
Image captioning has recently attracted great attention in the field of
artificial intelligence, due to the significant progress of machine
learning technologies and the release of a number of large-scale datasets
\cite{hossain2019comprehensive} \cite{bai2018survey} \cite{chen2017show}. 
The gist of the caption task is to generate a
meaningful and natural sentence that describes the most salient objects
and their interactions for the given image. Solving this problem has great
impact on human community, as it can help visual impaired people understand
various scenes and be treated as an auxiliary means of early childhood
education \cite{jiang2018recurrent} \cite{jiang2018learning}. Despite its 
widely practical applications, image captioning has
long been viewed as a challenging research, mainly because it needs to
explore a suitable alignment between two different modalities: image and
text.

The popular image captioning approaches adopt the encoder-decoder framework
\cite{vinyals2015show} \cite{jia2015guiding} \cite{wu2016encode}
\cite{mathews2016senticap} \cite{ramanishka2017top}. In general, a Convolutional Neural Network
(CNN) is often used as the encoder to present the image with a fixed-length
representation, while a Recurrent Neural Network (RNN) or Long Short-Term
Memory (LSTM) Network is employed to decode this representation into a
caption. Attention mechanism has indeed demonstrated significant effectiveness
on the task of image captioning \cite{xu2015show} \cite{you2016image}
\cite{anderson2018bottom} \cite{lu2017knowing} \cite{wang2017skeleton}. This mechanism allows models
to attend on image regions relevant to each generated word at every time
step, rather than only using the whole image to guide the generation of
captions. Although promising results have achieved, it’s obvious that the
current captioning systems are limited in two constraints:

First, since visual attention in captioning models can be viewed as the 
mappings from image regions to sentence snippets, however, this mapping 
procedure usually performs compulsively and unpredictably in a “black box”, 
and thus ignoring the fact that some words are not related to any entity 
in the image. As the result, it may cause inharmonious alignments between 
image regions and sentence snippets that will reduce the quality of generated 
sentences.

\begin{figure}[!t]
  \centering
  \includegraphics[width=3.5in]{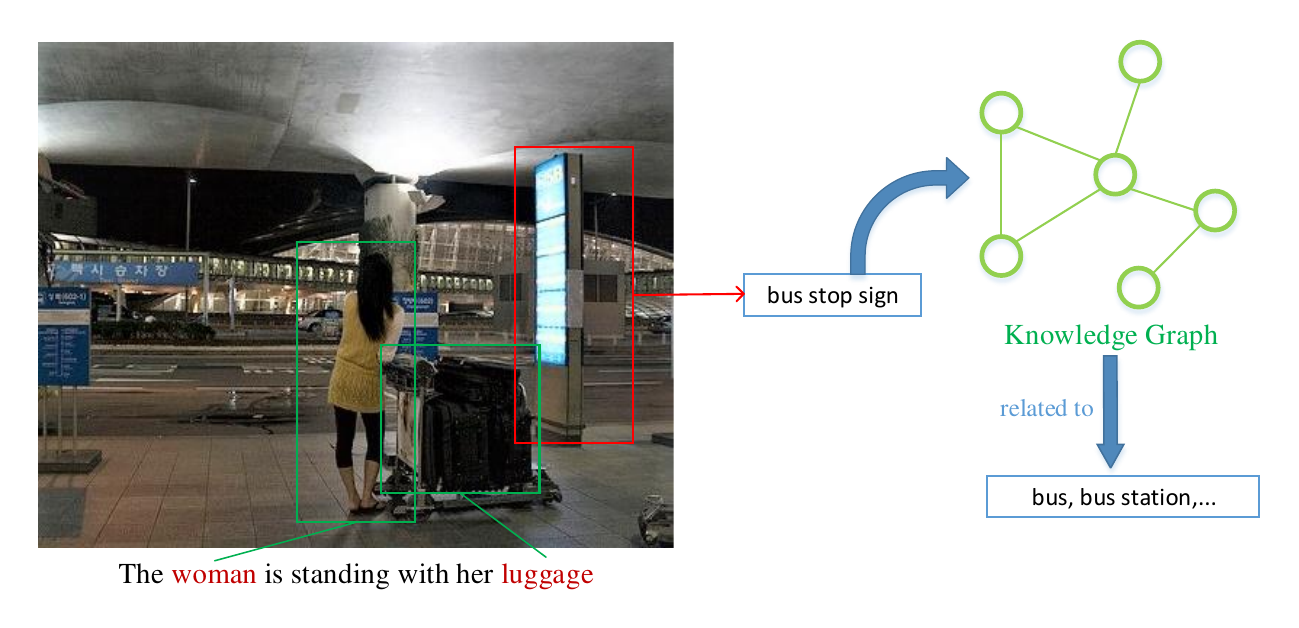}
%   where an .eps filename suffix will be assumed under latex,
%   and a .pdf suffix will be assumed for pdflatex; or what has been declared
%   via \DeclareGraphicsExtensions.
  \caption{The ground-truth caption simply describes the low-level content
  of the image, and doesn’t explain why this woman is standing there. By
  incorporating external knowledge, we speculated that she might be waiting
  for the bus. In the sentence, the words “woman” and “luggage” are more 
  important than others, as they describe the main aspects of the image.}
  \label{fig_sil}
\end{figure}

Second, most of the captioning models are built on a large number of paired
image-caption data, but each image in the training data only contains
several ground-truth captions, which will lack of sufficient cues to reveal
the incomprehensible intentions that are not explicitly presented in the
image. As shown in Fig.~\ref{fig_sil}. Furthermore, in order to extend 
the ability to describe new entities out of the training data, more 
knowledge need to be introduced from external data sources.

In this paper, we mainly focus on alleviating the aforementioned two
constraints to generate more accurate and meaningful captions. As we all
know, not all words in the caption have equivalent importance in describing
the image \cite{chen2017reference}. Seen from Fig.~\ref{fig_sil}. We capture
this perception, and devise a word attention to modulate the alignments
between words in sentence and regions in image. Specifically, a score is
assigned to each input word based on its significance in describing the
image, then we compute the word context vector at each time step to make
better use of the bottom-up semantic information to boost the process of
visual attention. This makes our model suit the human perception that some
salient regions in the image are more likely to be described than non-salient
ones. At the same time, we also leverage commonsense knowledge extracted
from knowledge graph to achieve better generalization.  In stead of fusing the
input sentence and external knowledge together to train an RNN, we input the knowledge
into the word generation stage to augment the probabilies of some potential words
that are likely to be applied to describe the given image.
This enables our
model to generate more meaningful sentences than other existing models.
To sum up, our contributions are shown as follows:
\begin{itemize}
  \item[$\bullet$] We propose a new text-dependent attention mechanism called word
  attention to assist the generation of visual attention, thereby making
  our model generate more accurate captions.
  \item[$\bullet$] We introduce a new strategy to incorporate knowledge 
  graph into our encoder-decoder framework to
  take better use of external knowledge to facilitate novel and meaningful
  captioning.
  \item[$\bullet$] By combining the aforementioned two proposed scenarios, experiments
  conducted on MSCOCO and Flickr30k benchmarks show that our approach
  achieves state-of-the-art performance and outperforms many of the
  existing approaches.
\end{itemize}

The rest of the paper is organized as: In section 2, we summarize the existing
captioning models into several categories, and review the previous works that
are related to this paper. Then, we present the implementation of our approach
in section 3. The experiment results and analysis are shown in section 4.
And in section 5, we make a conclusion and give a brief prospect of future work.

    % \begin{figure}[!t]
    %     \includegraphics[width=\textwidth]{fig1.eps}
    %     \caption{A figure caption is always placed below the illustration.
    %     Please note that short captions are centered, while long ones are
    %     justified by the macro package automatically.} \label{fig1}
    %     \end{figure}

\section{Related Work}
The state-of-the-art image captioning solutions are neural network-based
sequence learning methods, which use CNNs to encode an image into a 
fixed-length image representation, and then adopt RNNs to decode the 
representation into a meaningful sentence, thus the process of caption 
generation suits an end-to-end style.
Since there is a growing body of captioning algorithms have achieved 
superior performance \cite{vinyals2015show} \cite{jia2015guiding} 
\cite{mathews2016senticap}, these approaches often suffer from the problems 
of object missing and misprediction, due to their only use image-level 
representation to initialize the hidden state of RNN or LSTM.
\subsection{Attention mechanism based models}
\label{amm}
The problems of object missing and misprediction can be mitigated by 
introducing attention mechanism into the general captioning models. 
Motivated by  human perception, and encouraged by recent success in 
machine translation using attention mechanism, Xu et al. \cite{xu2015show} 
integrated visual attention into the encoder-decoder framework for image 
captioning, making their work a new state-of-the-art. Visual attention 
amounts to learning the latent alignments between words in sentence and 
regions in image when generating description word-by-word. Following this 
successful attempt, different attention methods are proposed. You et al. 
\cite{you2016image} proposed a semantic attention model which learns to 
selectively attend to semantic attributes detected from the given image, 
thereby making better use of the top-down visual information and the 
bottom-up semantic
information. Instead, Li et al. \cite{li2017image} combined the global
features and local features through a Global-Local attention, where the
local features are obtained by Faster R-CNN \cite{ren2015faster}. Similarly,
Anderson et al. \cite{anderson2018bottom} also implemented their bottom-up
attention using Faster R-CNN, and then a top-down attention is constructed
to attend to salient regions of the image.  Differently from the above
methods, Liu et al. \cite{liu2017attention} and Lu et al. \cite{lu2017knowing}
investigated the agreements between image regions and their corresponding
words. The former defined a quantitative metric to evaluate the “correctness”
of the attention maps generated by the uniform attention model and applied
supervision to improve the attention correctness, while the latter proposed
an adaptive attention and via a “visual sentinel” to decide when to attend
to the visual signals and when to depend on language properties to predict
the next word. 
In particular, Huang et al. \cite{huang2019attention} 
devised an AoA (Attention on Attention)
model to filter out the inappropriate attention results, and only used the
useful attended information to guide the caption generation process.
All of these works have demonstrated the effectiveness of
attention mechanism on the image captioning task. 
In this paper, a new text-dependent word attention is added to the uniform 
visual attention model. Its calculation only depends on the internal 
annotation knowledge in the training data, which can provide rich semantic 
information to guide the generation of visual attention.

\subsection{Incorporation of external knowledge}
\label{iek}
While promising advances are presented in exiting captioning methods, they
lack the ability to describe novel objects or attributes outside of
training corpora, and are unable to express implicit aspects of the image,
as the knowledge acquired from ground-truth captions is not sufficient.
Such issue can be solved by incorporating knowledge from external resources
into the caption generation process. An early study presented in
\cite{anne2016deep} exploited object knowledge from external object
recognition datasets or text corpora to facilitate novel object captioning.
Recently, Yao et al. \cite{yao2017incorporating} employed copying
mechanism to directly copy novel objects that do not exist in the training
corpora but are learnt from object recognition datasets to the output
sentence generated by LSTM, thus making their proposed model obtain
encouraging performance. Li et al. \cite{li2019pointing} further extended
this work by using pointing mechanism to elegantly accommodate the influence
between copying mechanism and word generation, in order to generate more
accurate and natural sentence. In contrast to using raw external data 
sources, some studies attempt to incorporate 
structured knowledge to solve the specific problems. Li et al. 
\cite{li2017incorporating}
 and Gu et al. \cite{gu2019scene} employed knowledge graph for visual 
 question answering and scene graph generation, 
respectively. These two studies both embedded the knowledge retrieved from external 
knowledge graph into a common space with other data, making their models flexible 
to adapt external test instances. 
Particularly, in \cite{zhou2019improving},
Zhou et al. proposed to leverage knowledge graph to boost image captioning,
which is close to our proposed approach. But unlike \cite{zhou2019improving}, 
which use a knowledge graph to extract indirectly related terms and directly 
related terms about the entities detected by an object detector to pretrain 
an RNN, we inject semantically related information of the detected objects 
into the output stage of the caption generator to augment the probability of 
some latent meaningful words at each decoding time. This also allows our 
system to generate more novel and meaningful captions.

\subsection{Enhanced by reinforcement learning}
\label{erl}
Moreover, several studies have been proposed to incorporate reinforcement
learning technology to address the issues of non-differentiable evaluation
metric and exposure bias \cite{ranzato2015sequence} in image captioning.
Ren et al. \cite{ren2017deep} treated image captioning as a decision-making
task, and proposed a deep reinforcement learning based model to generate a
natural description for an image. The model employs a “policy network” and
a “value network” to collaboratively determine the next word at each
intermediate step. In \cite{rennie2017self}, a self-critical sequence
training (SCST) approach is proposed to dramatically optimize the training
process using the test metrics (especially, the CIDEr metric) at decoding
stage. The SCST approach avoids having to estimate the reward signal and
consider how to normalize the reward, and it exploits the generated captions
in the inference phase as the “baseline” to encourage the generation of
more accurate captions. Anderson et al. \cite{anderson2018bottom} and
Qin et al. \cite{qin2019look} used the similar manner as \cite{rennie2017self}
to improve the performance of captioning models, but different in the way
of sequence sampling. It is worth mentioning that the last two works can
also be categorized as attention mechanism based method, since the former,
as mentioned above, proposed a bottom-up and top-down (Up-Down) attention
model, and the latter used this proposed model as the “backbone” to realize
its look back and predict forward (LBPF) model. 
Later on, Yao et al. \cite{yao2019hierarchy}
proposed to employ instance-level, region-level and image-level features
of an image to build a hierarchical structure, thereby giving the caption
generator a thorough image understanding. Their proposed architecture is
pluggable to many advanced reinforcement learning models, and encouraging
performance is achieved. In this work, we also follow these work, and 
use SCST approach to optimize our model.

\section{Method}
\label{met}
\begin{figure*}[!t]
  \centering
  \includegraphics[width=\textwidth]{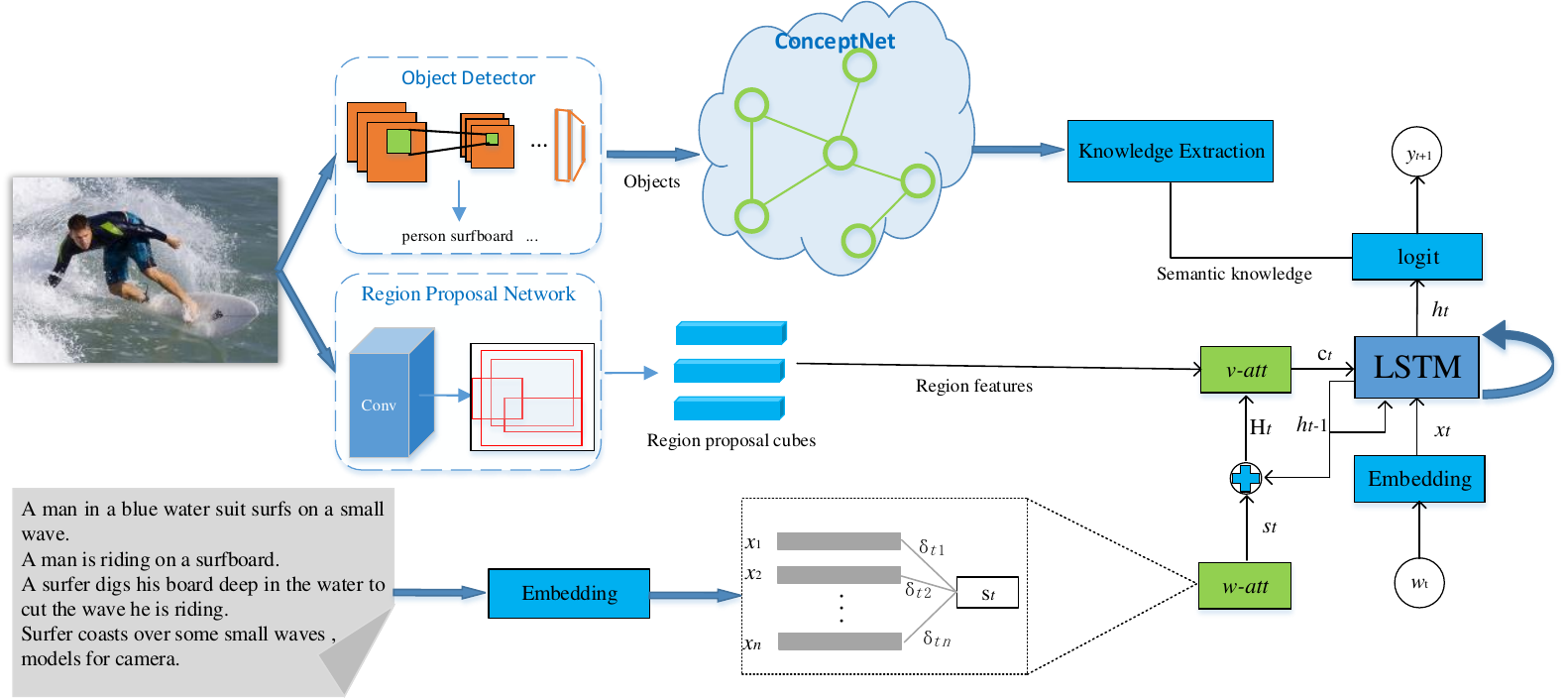}
  \caption{The overview of our captioning framework with word attention 
  and knowledge graph. Specifically, we use objects detected by an object 
  detector to retrieve semantic knowledge from the knowledge graph (here 
  we use ConceptNet) to guide the generation of captions. Meanwhile, to 
  extract more useful information from the given image, a region proposal 
  network is incorporated to generate the region features. Our proposed 
  word attention serves as another pipeline to inject compact textual 
  information to assist the calculation of visual attention. Here, 
  \emph{w-att} and \emph{v-att} represent word attention and visual 
  attention, respectively.}
  \label{fig_sio}
\end{figure*}

Like most of the captioning methods, we attempt to seed for a suitable and
human-like description for a given image. The overview of the proposed image
captioning architecture is illustrated in Fig.~\ref{fig_sio}. Compared 
to previous models, our model consists of two novel ideas. On the one hand, 
a special word attention is designed to handle the inharmonious matching 
problem between regions in image and words in caption. On the other hand, 
we take commonsense knowledge extracted from external knowledge graph into 
account to facilitate the
generation of novel and meaningful captions. Actually, our whole model seems
to make full use of the internal annotation knowledge and external knowledge
to guide the caption generation. But note that, the two proposed scenarios use knowledge in
different ways. In the following, we introduce the implementation of our whole model
including the extraction of image feature, the implementation of word attention,
the incorporation of knowledge graph and the usage of reinforcement learning.

\subsection{Image feature extraction and word embedding}
\label{ife}
To more effectively use the information in the image to guide the 
caption generation, we use region proposal features of the image to 
train our model. Specifically, following the method proposed by ren et 
al. \cite{ren2015faster}, we apply region proposal network to generate a 
lot of rectangular region proposals. Afterwards, each proposal is feed to 
ROI pooling layer and 3 full-connected layers to obtain a vector representation
$\boldsymbol{v}_{i}$ of each image region. Compared to the approach of Xu 
et al \cite{xu2015show}, which used the $z=x \times y$ locations of the 
activation grid to form the image representation, region proposal features 
provide more useful information of the image content. Please note that, 
given the image feature
$V=\left\{\boldsymbol{v}_{1}, \boldsymbol{v}_{2}, \dots, 
\boldsymbol{v}_{L}\right\}, \boldsymbol{v}_{i} \in \mathbb{R}^{D}$
, we use the mean pooling vector $\bar{\boldsymbol{v}}$ as the global 
image information, and the $\bar{\boldsymbol{v}}$ will be feed to 
initial the LSTM decoder to give an overall understanding of the image.

A common way of word embedding is one-hot encoding. This encoding style sets one
element of a vector to 1 and others to 0 to represent a specific word in the dictionary.
If there are too many words in the vocabulary, the one-hot vector will
become sparse and the problem of dimension explosion will also occur. Besides,
the one-hot encoding does not consider the order of words, which is unfavorable
to the calculation of our word attention. Thus, in this work, we use a pre-trained
word2vec model for word embedding. The word2vec is essentially a neural 
network, it uses the raw
one-hot vectors as inputs to obtain the final embedding vectors. Therefore, such
embedding manner fully considers the context between words and can 
provide more abundant semantic information.

\subsection{Implementation of word attention}
\label{iwa}
In this subsection, we introduce the implementation of the proposed word 
attention, and investigate in what manner the word attention contributes 
to the improvement of visual attention. Actually, the motivation of word 
attention comes from the perception that some words are more related to 
the content of a given image than others. What we need to do is to 
strengthen this connection, so that these words can play a better guiding 
role in the training process. Consequently, the
model can learn a more suitable mapping pattern between captions and 
images, which in turn improves the quality of generated captions.

Suppose an image \emph{I} to be described by a sentence
$S=\left\{w_{1}, w_{2}, \dots, w_{N}\right\}$,
where \emph{N} represents the length of the caption. The sequence learning
methods usually use RNN or LSTM to generate each word at each time stage,
in which LSTM has shown great performance. Following this trend, we add a
word attention into the caption generator to form our captioning model. At
the training phase, the word attention mainly depends on ground-truth captions,
and the operations of the word attention are as follows:
\begin{equation}\delta_{t i}=f_{w}\left({w}_{i}\right)\end{equation}
\begin{equation}\beta_{t i}=\frac{\exp \left(\delta_{t i}\right)}
  {\sum_{k=1}^{N} \exp \left(\delta_{t k}\right)}\end{equation}
\begin{equation}\boldsymbol{s}_{t}=\sum_{i=1}^{N} \beta_{t i} \boldsymbol{x}_{i}\end{equation}
where $f_{w}$ is a function that calculates the weight value allocated 
to ${w}_{i}$, $\boldsymbol{x}_{i}$ is the embedding vector of ${w}_{i}$,
  and $\boldsymbol{s}_{t}$ represents the word context vector at time \emph{t}.
Note that the $\delta_{t k}$  stays the same during the generation of each word
until the last time step. Here, inspired by the previous works \cite{kim2018distinctive}
\cite{chunseong2017attend}, we use TF-IDF method as the function $f_{w}$,
as this method can measure the importance degree of each word in a sentence
or document. The word context vector $\boldsymbol{s}_{t}$ is then fused
with the previous hidden state $\boldsymbol{h}_{t-1}$ of the LSTM decoder
to combine more compact semantic information to guide the visual attention,
calculated as follows:
\begin{equation}\boldsymbol{H}_{t}=\boldsymbol{s}_{t} \odot 
  \boldsymbol{h}_{t-1}\end{equation}
\begin{equation}e_{t i}=\boldsymbol{W}_{e}^{T} \tanh \left(\boldsymbol{W}_{v} 
  \boldsymbol{v}_{i}+\boldsymbol{W}_{h} \boldsymbol{H}_{t}\right)\end{equation}
\begin{equation}\alpha_{t i}=\frac{\exp \left(e_{t i}\right)}{\sum_{k=1}^{L} 
  \exp \left(e_{t k}\right)}\end{equation}
\begin{equation}\boldsymbol{c}_{t}=\sum_{i=1}^{L} \alpha_{t i} 
  \boldsymbol{v}_{i}\end{equation}
where $\odot$ is the element-wise multiplication; $\boldsymbol{W}_{e}$, 
$\boldsymbol{W}_{v}$ and $\boldsymbol{W}_{h}$ are learned parameters; 
\emph{tanh} is the hyperbolic tangent function; the visual context vector
$\boldsymbol{c}_{t}$ is the weighted sum of all the image region features. 
Combined with word attention, the decoder LSTM updates for time step 
\emph{t} are:
\begin{equation}\boldsymbol{i}_{t}=\sigma\left(\boldsymbol{W}_{i} \boldsymbol{x}_{t}+
  \boldsymbol{U}_{i} \boldsymbol{c}_{t}+\boldsymbol{Z}_{i} \boldsymbol{h}_{t-1}+
  \boldsymbol{b}_{i}\right)\end{equation}
\begin{equation}\boldsymbol{f}_{t}=\sigma\left(\boldsymbol{W}_{f} 
  \boldsymbol{x}_{t}+\boldsymbol{U}_{f} \boldsymbol{c}_{t}+
  \boldsymbol{Z}_{f} \boldsymbol{h}_{t-1}+\boldsymbol{b}_{f}\right)\end{equation}
\begin{equation}\boldsymbol{o}_{t}=\sigma\left(\boldsymbol{W}_{o} 
  \boldsymbol{x}_{t}+\boldsymbol{U}_{o} \boldsymbol{c}_{t}+
  \boldsymbol{Z}_{o} \boldsymbol{h}_{t-1}+\boldsymbol{b}_{o}\right)\end{equation}
\begin{equation}\boldsymbol{g}_{t}=\sigma\left(\boldsymbol{W}_{g} 
  \boldsymbol{x}_{t}+\boldsymbol{U}_{g} \boldsymbol{c}_{t}+
  \boldsymbol{Z}_{g} \boldsymbol{h}_{t-1}+\boldsymbol{b}_{g}\right)\end{equation}
\begin{equation}\boldsymbol{m}_{t}=\boldsymbol{f}_{t} \odot 
  \boldsymbol{m}_{t-1} + \boldsymbol{i}_{t} \odot \boldsymbol{g}_{t}\end{equation}
\begin{equation}\boldsymbol{h}_{t}=\boldsymbol{o}_{t} \odot \tanh 
  \left(\boldsymbol{m}_{t}\right)\end{equation}
\begin{equation}p_{t+1}=\boldsymbol{w}_{t+1}^{T} \boldsymbol{M}_{g} 
  \boldsymbol{h}_{t}\end{equation}
here $\boldsymbol{i}_{t}$, $\boldsymbol{f}_{t}$, $\boldsymbol{o}_{t}$, 
$\boldsymbol{g}_{t}$, $\boldsymbol{m}_{t}$ and $\boldsymbol{h}_{t}$
are are input gate, forget gate, output gate, cell gate, cell memory 
and hidden state of the LSTM, respectively;
$\sigma(.)$ represents the sigmoid function; $\boldsymbol{W}_{\ast}$, 
$\boldsymbol{U}_{\ast}$, $\boldsymbol{Z}_{\ast}$, $\boldsymbol{b}_{\ast}$ 
are weight matrices and biases to be learned; equation (14) adopts the 
generation mechanism to predict the next word, where $\boldsymbol{M}_{g}$ 
is the transformation matrix.

Finally, our proposed word attention can be insert into the encoder-decoder 
framework in a trainable manner, thus making better use of the caption 
information annotated by humans. As a result, the quality of generated 
captions will be improved.

\subsection{Incorporation of knowledge graph}
\label{ikg}
On the task of image captioning, knowledge is of significantly important, 
as it provides a lot of cues for generating captions. The ground-truth 
annotations corresponding to each image in paired image-caption datasets 
are the knowledge provided by human beings for caption generation, which 
can be called internal knowledge. However, in many existing datasets, it 
is impossible to include all the knowledge required for captioning task, 
thereby limiting research progress. Therefore, we acquire knowledge from 
external resources to assist the caption generation, so as to improve the 
generalization performance of the captioning model. In recent years, many 
knowledge graph have appeared in the field of artificial intelligence. In 
this paper, we use ConceptNet \cite{speer2017conceptnet}, which is an open 
multilingual knowledge graph containing common sense knowledge closely 
related to human daily life, to help computers understand human intentions.

\begin{figure}[!t]
  \centering
  \includegraphics[width=3.8in]{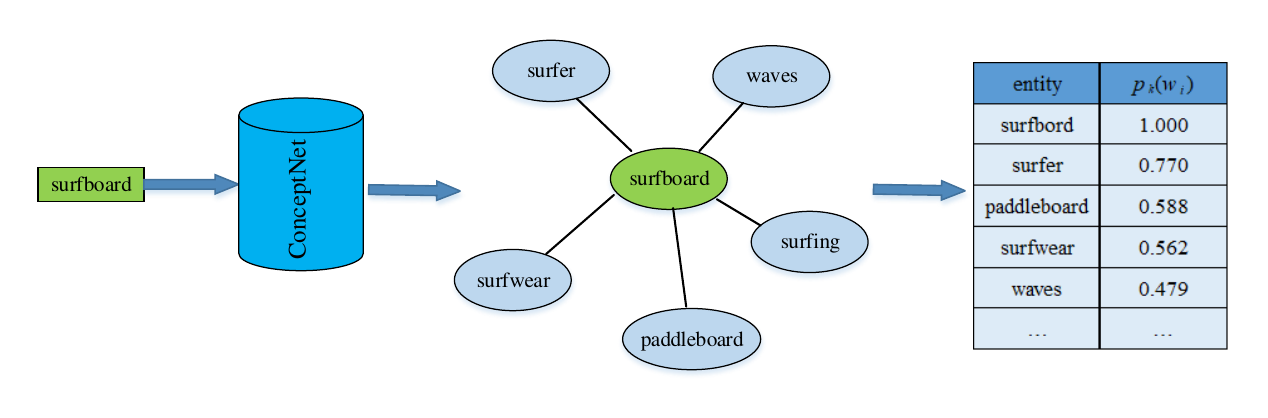}
  \caption{The illustration of knowledge extraction using an object “surfboard”.
  For convenience, we simply give a part of the results. Note that each
  relevant semantic entity corresponds to a probability $p_{k}$ that represents
  the degree of correlation.}
  \label{fig_sit}
\end{figure}

In general, each piece of knowledge in the knowledge graph can be view as
a tripe (\emph{subject}, \emph{rel}, \emph{object}), where \emph{subject}
and \emph{object} represent two entities or concepts in the real world,
and \emph{rel} is the relationship between them. To obtain informative
knowledge that are relevant to the given image, we first use Faster R-CNN \cite{ren2015faster}
to detect a series of objects or visual concepts, and then use these objects
or concepts to retrieve semantically similar knowledge from the knowledge
graph. Fig.~\ref{fig_sit} gives an illustration of using a detected word
"\emph{surfboard}" to retrieve sematic knowledge from the ConceptNet. As we
have seen, each knowledge entity corresponds to a probability $p_{k}$ that
represents the degree of correlation. For each detected object or concept,
we select the relevant knowledge entities for captioning task. Such
we get a small semantic knowledge corpus $W_{k}$ containing the most
relevant knowledge.

And the problem is that how to apply the important semantic knowledge 
extracted from the knowledge graph to guide
the process of caption generation, which needs to be carefully considered.
Since unnecessary inputs may cause noise in the training phase, thus
reducing the performance of the model. 
Thereby, we do not directly input
the semantic knowledge into the input layer of LSTM for training, instead
we focus on changing the \emph{logit} layer of the caption generation
network, and we augment the probability of some potential words that are appear in the
constructed semantic knowledge corpus $W_{k}$ when we predict the next word. After
using Back Propagation to train the whole model, a more robust system is obtained.
For this purpose, we changed the equation (14) as:
\begin{equation}p_{t+1}=\left\{\begin{array}{l}
  \boldsymbol{w}_{t+1}^{T} \boldsymbol{M}_{g} \boldsymbol{h}_{t}+\lambda p_{k}
  \left(\boldsymbol{w}_{t+1}\right),  \boldsymbol{w}_{t+1} \in \boldsymbol{W}_{K} \\
  \boldsymbol{w}_{t+1}^{T} \boldsymbol{M}_{g} \boldsymbol{h}_{t},  otherwise
  \end{array}\right.\end{equation}

where $\lambda$  is a hyper parameter that measures the degree of introducing
external semantic knowledge.
% $p_{k}\left(\boldsymbol{w}_{t+1}\right)$ will
% be set to zero if the word $\boldsymbol{w}_{t+1}$ not exist in $W_{k}$,
% otherwise it represents the probability given in the knowledge retrieval
% process.
If the word $\boldsymbol{w}_{t+1}$ exists in the constructed semantic knowledge corpus $W_{k}$,
the prediction probability of the word will be determined by the prediction
probability of the generation mechanism and the corresponding retrieval
probability $p_{k}\left(\boldsymbol{w}_{t+1}\right)$.
In general, a softmax function is used after equation (15) to
obtain a normalized word probability distribution. By adding an additional
probability to each possible word, the model will be able to discover some
implicit cues, thus leading to generate more novel and meaningful captions.

\subsection{Sequence generation based on reinforcement learning}
\label{sgr}
Here we discuss the caption generation process of our model. It is sure that
in the training and inference phase, our proposed two scenarios can not only
be appropriately combined together to guide the caption generation process,
but each can perform solely to address the different deficiencies existing
in the previous models. Our approach suits the popular sequence learning based
methods. In other word, the sequence is generated word-by-word.
The state-of-the-art captioning models are typically trained with 
cross entropy loss:
\begin{equation}L(\theta)=-\sum_{t=1}^{T} \log \left(p_{\theta}
  \left(w_{t}^{*} | w_{1}^{*}, \ldots, w_{t-1}^{*}\right)\right)\end{equation}
where $p_{\theta}$ is a captioning model with all the parameters 
$\theta$, and the sequence $\left(w_{1}^{*}, \ldots, w_{T}^{*}\right)$ 
is the ground-truth caption.
However, the cross entropy objective function dose not perform well with the
problem of "exposure bias" \cite{ranzato2015sequence}. The issue can be mitigated
by introducing "scheduled sampling" \cite{bengio2015scheduled} at the decoding stage.
However, the "scheduled sampling" strategy seems to be statistically inconsistent.
Another alternative solution to "exposure bias" is reinforcement learning 
\cite{liu2016optimization} \cite{liu2017improved}.
In this paper, we adopt SCST method to 
optimize our model. 
Note that, the caption generator (LSTM) can be viewed as an "agent", and 
the caption words and image features serve as “environment”. In addition,
$p_{\theta}$ defines a "policy", that will result in generating the next 
best word.
In this case, we minimize the negative expected reward to train the model:
\begin{equation}L_{r}(\theta)=-\mathbb{E}_{w_{1: T}  p_{\theta}}
  \left[r\left(w_{1: T}\right)\right]\end{equation}
where $r\left(w_{1: T}\right)$ is the score function, here we use CIDEr 
as \cite{rennie2017self} and \cite{anderson2018bottom}. The gradient of 
this loss then will be approximated as:
\begin{equation}\nabla_{\theta} L_{r}(\theta) \approx-\left(r\left(w_{1: T}^{s}
  \right)-r\left(w_{1: T}^{m}\right)\right) \nabla_{\theta} \log p_{\theta}
  \left(w_{1: T}^{s}\right)\end{equation}
where $w_{1: T}^{s}$ and $w_{1: T}^{m}$ denote the sampled sequence and 
the result of greedy decoding, respectively. In particular, we set the 
baseline as $r\left(w_{1: T}^{m}\right)$, this has made our model achieve
significant gains in the performance.

The core idea of this reinforcement learning based training approach is to take
the reward obtained by the inference algorithm used by the current model during
testing as the baseline of the reinforce algorithm.  This approach keeps the model
consistent during training and inference, thereby significantly improving the quality of
the generated captions. In the later, we demonstrate the effectiveness of our model
combining with reinforcement learning based training manner.

\section{Experiments}
\label{exp}
In this section, extensive experiments are conducted to evaluate the effectiveness of 
our proposed model. We first introduce the datasets and evaluation metrics. Then, the 
implementation details of our experiments are followed. Finally, we make a 
comparison with other state-of-the-art models and give brief analysis of 
the results.

\subsection{Datasets and evaluation metrics}
\label{dem}
We mainly use the popular MSCOCO 2014 dataset to validate the performance 
of our proposed models. This large dataset contains 123,287 images, with 
at least 5 ground-truth sentences are annotated per image for image captioning
%(each of which is annotated with 5 captions).
For fair comparison with other methods, we adopt the ‘Karpathy’ splits 
\cite{johnson2016densecap} that have been widely used in previous work. 
Thus, we get 113,287 images for training, and 5,000 images for validation 
and testing respectively.
Compared to MSCOCO 2014 dataset, flickr30k dataset is smaller, and it 
contains 31,000 images. Since it does not provide official split, we also 
follow the split presented in work \cite{johnson2016densecap}, i.e., 29,000 
images for training, 1,000 images for validation, and 1,000 images for testing.
In the training phase, we select the most 8000 common words in the COCO 
captions to build our vocabulary, and each word in the vocabulary is 
presented as a 512-demensional vector.

To automatically evaluate the quality of machine-generated captions remains
a great challenge, mainly because of the fact that machines lack of the
ability to make a suitable judgment independently as humans. Most of the
existing evaluation metrics for image captioning task attempt to calculate
a quantitative value according to the consistency of ground-truth
annotations and generated captions. Here, we briefly introduce the
evaluation metrics used in the experiment, including BLEU \cite{papineni2002bleu}, 
METEOR \cite{banerjee2005meteor}, CIDEr-D \cite{vedantam2015cider}
and ROUGE-L \cite{lin2003automatic}. BLEU is the most common matric for the evaluation of machine
generated sentence. Since BLEU is based on n-gram precision, we choose
BLEU-1, BLEU-2, BLEU-3 and BLEU-4 to evaluate the performance. METEOR is 
designed to make up for the deficiency of BLUE metric, and it takes 
sentence stems and synonyms into account to evaluate the generated 
sentence. CIDEr-D metric aims to measure the consensus between human 
annotations and generated captions, it is mainly designed for 
automatically image captioning. In addition, ROUGE-L metric pays more 
attention on recall rate, and is also employed to our experiments.

\subsection{Implementation details}
\label{ide}
In this work, we use Faster R-CNN as the object detector to detect a number
of objects. The Faster R-CNN is pretrained on the Visual Genome dataset, and then
fine-tuned on the MSCOCO dataset. The RPN network, which is part of the
Faster R-CNN, is also used to generate the region features of the given image.
As a result, for each $256 \times 256$ size image, We get 36 2048-dimensional image region vectors.
In the decoding stage, we use the
LSTM network as the decoder, and the input and hidden layers are both set to
512. The dimension of word embedding is also set to 512. Since complex network structure usually leads to overfitting,
we adopt dropout method to randomly inactivate some neurons, here the
dropout rate is set to 0.5. The hyper parameter $\lambda$ is
empirically set to 0.2, and we will discuss the selection of $\lambda$ in
the later.

 The training process is divided into the following two stages: In the first stage,
the model is trained under cross-entropy, and the mini-batch size is set to 64.
In particular, we use the Adam optimization function to optimize our network,
with the initial learning rate of $5 \times 10^{-4}$ and the momentum of 0.9.
At every 5 epochs, the learning rate is annealed by 0.7. In order to 
obtain a more generalized model,
we select the BLEU-4 metric to monitor the training process. Early stopping is applied
when the BLEU-4 score continues decrease in 5 consecutive epochs, and the
maximum iteration is set to 30 epochs.
In the second stage, the reinforcement
learning optimization algorithm is run to further optimize the model. At
this stage, the training epoch is set to 20, the training batch size is 
adjusted to 32, the learning rate is fixed at $1 \times 10^{-4}$, and
other parameters remain unchanged. During the inference phase, we use beam search technology to
select the most appropriate caption from candidate captions, and the
beam size is set to 3. The maximum length of generated sentence here is set to 16.

\subsection{Experimental results and analysis}
\label{era}
\subsubsection{Results on MSCOCO and flick30k datasets}
\label{rmf}
As mentioned above,
we empirically evaluate the effectiveness of our model on MSCOCO and
flickr30k datasets. In the following, we compare our model with other
state-of-the-art models, including attention based models, knowledge
incorporated models and reinforcement learning Enhanced models.

Table ~\ref{table_one} shows the comparison results of our proposed
model with other state-of-the-art models on MSCOCO dataset.
Except for the UP-Down model, our model outperforms
all the compared models.
 Even compared with the state-of-the-art Up-Down model, our model obtains 
 superior results on several
metrics, especially for BLEU, ROUGE-L and CIDEr-D metrics, we achieve BLEU-2 / BLEU-3
/ BLEU-4 / ROUGE-L /CIDEr-D scores of 0.638, 0.490, 0.373, 0.574 and 1.212, respectively.
The Up-Down model uses bottom-up image features acquired by Faster R-CNN 
and CIDEr-D optimization technology to train 
the network, which is similar to ours. Besides, this model especially uses 
two LSTMs to generate captions, which can increase the gains to some extent. 
In contrast, we only employ one LSTM at the decoding stage, this 
may reduce the performance of the model. 
However, by incorporating word attention and knowledge graph, our full model 
can achieve comparable results even better results than the UP-Down model.
% These are the main reasons why our full model works better on some metrics 
% but works worse on another metrics than Up-Down model.

 \begin{table*}[!t]
  \renewcommand{\arraystretch}{1.0}
   \renewcommand\tabcolsep{2.0pt}
  %if using array.sty, it might be a good idea to tweak the value of
  %\extrarowheight as needed to properly center the text within the cells
  \caption{Performance comparison with other state-of-the-art models on the MSCOCO benchmark}
  \label{table_one}
  \centering
  % Some packages, such as MDW tools, offer better commands for making tables
  % than the plain LaTeX2e tabular which is used here.
  \begin{tabular}{l|cccccccc}
  \toprule
  Method  &  BLEU-1  &  BLEU-2  &  BLEU-3  &  BLEU-4  &  METEOR  &  ROUGE-L  &  CIDEr-D\\
  \hline
  Google NIC\cite{vinyals2015show} & 0.666 & 0.461 & 0.329 & 0.246 & - & - &  -\\
  Hard-Att\cite{xu2015show} & 0.718 & 0.504 & 0.357 & 0.250 & 0.230  & - &  -\\
  ATT\cite{you2016image} & 0.731 & 0.565 & 0.424 & 0.316 & 0.250 & 0.535 & 0.943\\
  Review Net\cite{wu2016encode} & 0.720 & 0.550 & 0.414 & 0.313 & 0.256 & 0.533 & 0.965 \\
  Areas-Att\cite{pedersoli2017areas} & - & - & - & 0.307 & 0.245 & - & 0.938\\
  Adaptive\cite{lu2017knowing} & 0.742 & 0.580 & 0.439 & 0.332 & 0.266 & - & 1.085\\
  Saliency-Att\cite{cornia2018paying} & 0.708 & 0.536 & 0.391 & 0.284 & 0.248 & 0.521 & 0.898\\
  CNet-NIC\cite{zhou2019improving} & 0.731 & 0.549 & 0.405 & 0.299 & 0.256 & 0.539 & 1.072\\
  SCST:Att2in\cite{rennie2017self} & - & - & - & 0.333 & 0.263 & 0.553 & 1.114\\
  Up-Down\cite{anderson2018bottom} & \textbf{0.798} & - & - & 0.363 & \textbf{0.277} & 0.569 & 1.204\\
  \hline
  ours & 0.793 & \textbf{0.638} & \textbf{0.490} & \textbf{0.373} & 0.273 &
  \textbf{0.574} & \textbf{1.212}\\

  \bottomrule
  \end{tabular}
\end{table*}

\begin{table*}[!t]
  \renewcommand{\arraystretch}{1.0}
   \renewcommand\tabcolsep{2.4pt}
  %if using array.sty, it might be a good idea to tweak the value of
  %\extrarowheight as needed to properly center the text within the cells
  \caption{Quantitative results on the use of different components.}
  \label{table_two}
  \centering
  % Some packages, such as MDW tools, offer better commands for making tables
  % than the plain LaTeX2e tabular which is used here.
  \begin{tabular}{ccc|cccccccc}
  \toprule
  RL & WA & KG &  BLEU-1  &  BLEU-2  &  BLEU-3  &  BLEU-4  &  METEOR  &  ROUGE-L  &  CIDEr-D\\
  \hline
 
  $\surd$ &  &  &  0.773 & 0.620 & 0.470 & 0.364 & 0.270 & 0.561 & 1.132\\
  $\surd$ & $\surd$ &  &  0.775 & 0.618 & 0.476 & 0.358 & 0.269 & 0.566 & 1.165\\
  $\surd$ &  & $\surd$ &  0.784 & 0.627 & 0.482 & 0.363 & 0.270 & 0.563 & 1.178\\
  $\surd$ & $\surd$ & $\surd$ &  \textbf{0.793} & \textbf{0.638} & 
  \textbf{0.490} & \textbf{0.373} & \textbf{0.273} &
  \textbf{0.574} & \textbf{1.212}\\

  \bottomrule
  \end{tabular}
\end{table*}

To evaluate the effectiveness of our each design, we implement our several models
with different components. Let RL denotes our reinforcement learning baseline model,
RL+WA only incorporates word attention to boost captioning, RL+KG only
introduces the knowledge graph into the baseline architecture,
and RL+MA+KG represents the full model that combined with word attention and knowledge
graph.
We can see from Table ~\ref{table_two}, the RL+WA model and RL+KG model 
show better results over several metrics than the baseline 
model (RL), and our RL+WA+KG model achieves best performance compared to other 
models. The results demonstrate that not only the 
performance of the model can be improved by using word attention and 
knowledge graph alone, but also the combination of these two designs can 
make the model get better performance. In addition, the use of knowledge 
graph can bring more benefits than word attention, indicating that through the 
incorporation of external knowledge, our model is able to discover more 
important cues to describe the given image. 

Table ~\ref{table_three} shows the comparison results on Flickr30K 
dataset. By incorporating reinforcement learning technology
to optimize our model, it is not surprising that we outperform all the
compared models. The flickr30k dataset contains less training data, Thereby
leading to a small reduction  in the performance. Even that we achieve superior
performance on all the standard evaluation metrics.
Especially, we show the comparison results on the online MSCOCO evaluation server.
As shown in Table ~\ref{table_four}.
Compared to the state-of-the-art models, we also achieve better results on
almost all the metrics.

\begin{table*}[!t]
  \renewcommand{\arraystretch}{1.0}
   \renewcommand\tabcolsep{2.0pt}
  %if using array.sty, it might be a good idea to tweak the value of
  %\extrarowheight as needed to properly center the text within the cells
  \caption{performance comparison with other state-of-the-art models on the flickr30k benchmark}
  \label{table_three}
  \centering
  % Some packages, such as MDW tools, offer better commands for making tables
  % than the plain LaTeX2e tabular which is used here.
  \begin{tabular}{l|cccccccc}
  \toprule
  Method  &  BLEU-1  &  BLEU-2  &  BLEU-3  &  BLEU-4  &  METEOR  &  ROUGE-L  &  CIDEr-D\\
  \hline
  Google NIC\cite{vinyals2015show} & 0.663 & 0.423 & 0.277 & 0.183 & - & - &  -\\
  Hard-Att\cite{xu2015show} & 0.669 & 0.439 & 0.296 & 0.199 & 0.185  & - &  -\\
  ATT\cite{you2016image} &  0.647 & 0.460 & 0.324 & 0.230 & 0.189 & - & -\\
  SCA-CNN\cite{chen2017sca} & 0.662 & 0.468 & 0.325 & 0.223 & 0.195 & - & -\\
  Adaptive\cite{lu2017knowing} & 0.677 & 0.494 & 0.354 & 0.251 & 0.204 & - & 0.531 \\
  Att-Region\cite{wu2017image} & 0.730 & 0.550 & 0.400 & 0.280 & - & - & -\\
  \hline
  ours & \textbf{0.745} & \textbf{0.566} & \textbf{0.417} & \textbf{0.313} 
  & \textbf{0.221} & \textbf{0.498} & \textbf{0.716}\\
 \bottomrule
  \end{tabular}
\end{table*}

\begin{table*}[!t]
  \renewcommand{\arraystretch}{1.0}
   \renewcommand\tabcolsep{2.0pt}
  %if using array.sty, it might be a good idea to tweak the value of
  %\extrarowheight as needed to properly center the text within the cells
  \caption{Performance comparison with other state-of-the-art models on the online MSCOCO server.}
  \label{table_four}
  \centering
  % Some packages, such as MDW tools, offer better commands for making tables
  % than the plain LaTeX2e tabular which is used here.
  \begin{tabular}{l|cccccccccccccccc}
  \toprule
  & \multicolumn{2}{c}{\underline{BLEU-1}}   &
  %\multicolumn{2}{c}{\underline{BLEU-2}}  &  \multicolumn{2}{c}{\underline{BLEU-3}}  &
  \multicolumn{2}{c}{\underline{BLEU-4}}  &  \multicolumn{2}{c}{\underline{METEOR}}  
  &  \multicolumn{2}{c}{\underline{ROUGE-L}}  &  \multicolumn{2}{c}{\underline{CIDEr-D}} 
  \\
  %c5 & c40\\
  Method  & c5 & c40  &
  %c5 & c40  &   c5 & c40  &
  c5 & c40  &   c5 & c40  &   c5 & c40  &  c5 & c40\\
  \hline
  Google NIC\cite{vinyals2015show} & 0.713 & 0.895
  %& 0.542 & 0.802 & 0.407 & 0.694
  & 0.309 & 0.587 & 0.254 & 0.346 & 0.530 & 0.682 & 0.943 & 0.946\\
  Hard-Att\cite{xu2015show} & 0.705 & 0.881
 % & 0.528 & 0.779 & 0.383 & 0.658
  & 0.277 & 0.537 & 0.241 & 0.322 & 0.516 & 0.654 & 0.865 & 0.893 \\
  ATT\cite{you2016image} & 0.731 & 0.900 &
  %0.565 & 0.815 & 0.424 & 0.709 &
  0.316 & 0.599 & 0.250 & 0.335 & 0.535 & 0.682 & 0.943 & 0.958 \\
  Review Net\cite{wu2016encode} & 0.720 & 0.900 &
  %0.550 & 0.812 & 0.414 & 0.705 &
  0.313 & 0.597 & 0.256 & 0.347 & 0.533 & 0.686 & 0.965 & 0.969  \\
  MSM \cite{yao2017boosting} & 0.739 & 0.919 &
  %0.575 & 0.842 & 0.436 & 0.740 &
  0.330 & 0.632 & 0.256 & 0.350 & 0.542 & 0.700 & 0.984 & 1.003 \\
  % Saliency-Att\cite{cornia2018paying} & 0.708 & 0.536 & 0.391 & 0.284 & 0.248 & 0.521 & 0.898\\
  % CNet-NIC\cite{zhou2019improving} & 0.731 & 0.549 & 0.405 & 0.299 & 0.256 & 0.539 & 1.072\\
  Adaptive\cite{lu2017knowing} & 0.748 & 0.920 & 0.336 & 0.637 & 
  0.264 & 0.359 & 0.550 & 0.705 & 1.042 & 1.059 \\
  SCST:Att2in\cite{rennie2017self} & 0.781 & 0.937 &
  %0.619 & 0.860 & 0.470 & 0.759 &
  0.352 & 0.645 & 0.270 & 0.355 & 0.563 & 0.707 & 1.147 & 1.167\\
  Up-Down\cite{anderson2018bottom} & \textbf{0.802} & \textbf{0.952} &
  %\textbf{0.641} & 0.888 & \textbf{0.491} & 0.794 &
  0.369 & 0.685 & 0.276 & 0.367 & 0.571 & 0.724 & 1.179 & 1.205\\
  \hline
  ours &  0.798 & 0.950 &
  %0.631 & \textbf{0.895} & 0.488 & \textbf{0.805} &
  \textbf{0.376} & \textbf{0.688} & \textbf{0.284} & \textbf{0.376} & 
  \textbf{0.580} & \textbf{0.730} & \textbf{1.218} & \textbf{1.229}\\
  \bottomrule
  \end{tabular}
\end{table*}

\subsubsection{The analysis of introducing external knowledge }
\label{aiek}
\begin{figure}[!tbp]
  \centering
  \includegraphics[width=4.0in]{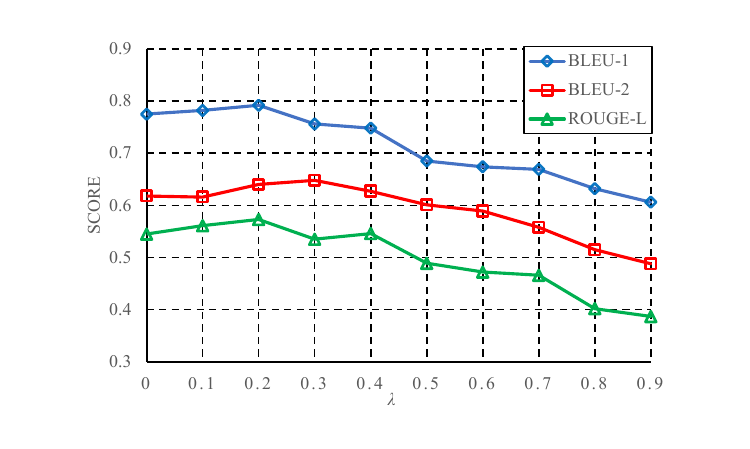}
  \caption{The change of BLEU-1, BLEU-2 and ROUGE-L scores after selecting 
  diferent $\lambda$ values. This experiment is conducted on MSCOCO benchmark.}
  \label{fig_sif}
\end{figure}

Acquiring the knowledge from the previous work \cite{li2019learning}, we
find that each image in the captioning datasets (e.g. MSCOCO, Flickr30k)
usually contains a small number of objects.
% , the most common case, varies
% from 1 to 4.
Therefore, in the experiment, we select the top-3 objects
according to the detected score, and inject them into the ConceptNet to
retrieve semantic knowledge to boost the caption generation. Hence, 
given an image we get a series of relevant knowledge, with each piece 
of knowledge contains two parts: semantic entity and the relevant 
probability. Then, we consider to use $\lambda$ to control the degree of
introducing external knowledge. We intuitively choose the value of $\lambda$ from
0 to 0.9. Fig.~\ref{fig_sif} shows the change of BLEU-1, BLEU-2
and ROUGE-L scores conditioned on the selection of parameter $\lambda$. We can see that
When $\lambda=0.2$ the BLEU-1 score and ROUGE-L score reach the peak 
simultaneously, while the BLEU-2 score reach its maximum at $\lambda=0.3$.
With the $\lambda$ value continues to increase, the scores of these 
three metrics gradually decrease. We speculate that the large
$\lambda$ may reduce the stability of the training model, while a smaller
$\lambda$ provides little benefits to the caption generation process.
To bring a bigger benefit, we set the $\lambda$ value to 0.2 in other experiments.

\begin{figure*}[!htp]
  \centering
  \subfigure[Model with word attention]{
  \begin{minipage}{9cm}
    \centering
    \includegraphics[width=4.0in]{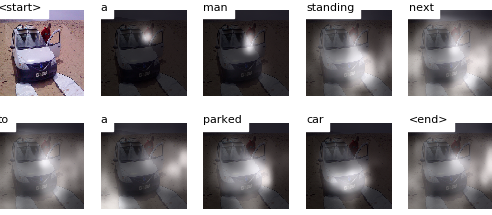}
  \end{minipage}
  }
  \subfigure[Model without word attention]{
  \begin{minipage}{9cm}
    \centering
    \includegraphics[width=4.0in]{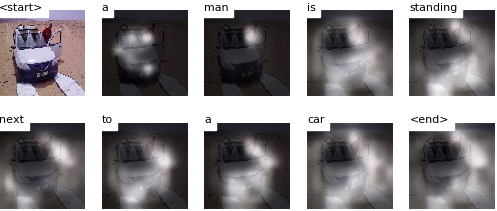}
  \end{minipage}
  }
    \caption{An example illustrating the effectiveness of word attention. 
    $<$ start $>$ and $<$ end $>$ are tokens representing the beginning and
    end of sentence respectively.}
    \label{fig_sip}
\end{figure*}

\subsubsection{Attention analysis}
\label{voi}
% As the previous work \cite{xu2015show}, we visualize the attention on individual pixels, 
% thereby better reveal the correctness of the visual attention.
As the previous work \cite{xu2015show}, we visualize the attention on individual pixels, 
thereby better revealing the correctness of the visual attention.
Fig.~\ref{fig_sip} shows an example of word generation when attending to 
different regions of the image. (a) and (b) are models with and without word 
attention respectively. As indicated by this example, when generating the descriptive 
words, especially the more important words, like "man" and "car", 
the model with word attention can more accurately focus on the appropriate 
positions of the image. Compared to the model without word attention, 
the model with word attention 
can predict the word "parked", indicating that more fine-gained captions 
can be generated by our proposed model.
%they are corresponded to the correct image regions.
The result shows that our proposed word attention 
combining with the standard visual attention can perform well with the word generation process, and 
facilitate to generate more accurate and fine-grained captions.

\begin{figure*}[!t]
  \centering
  \includegraphics[width=\textwidth]{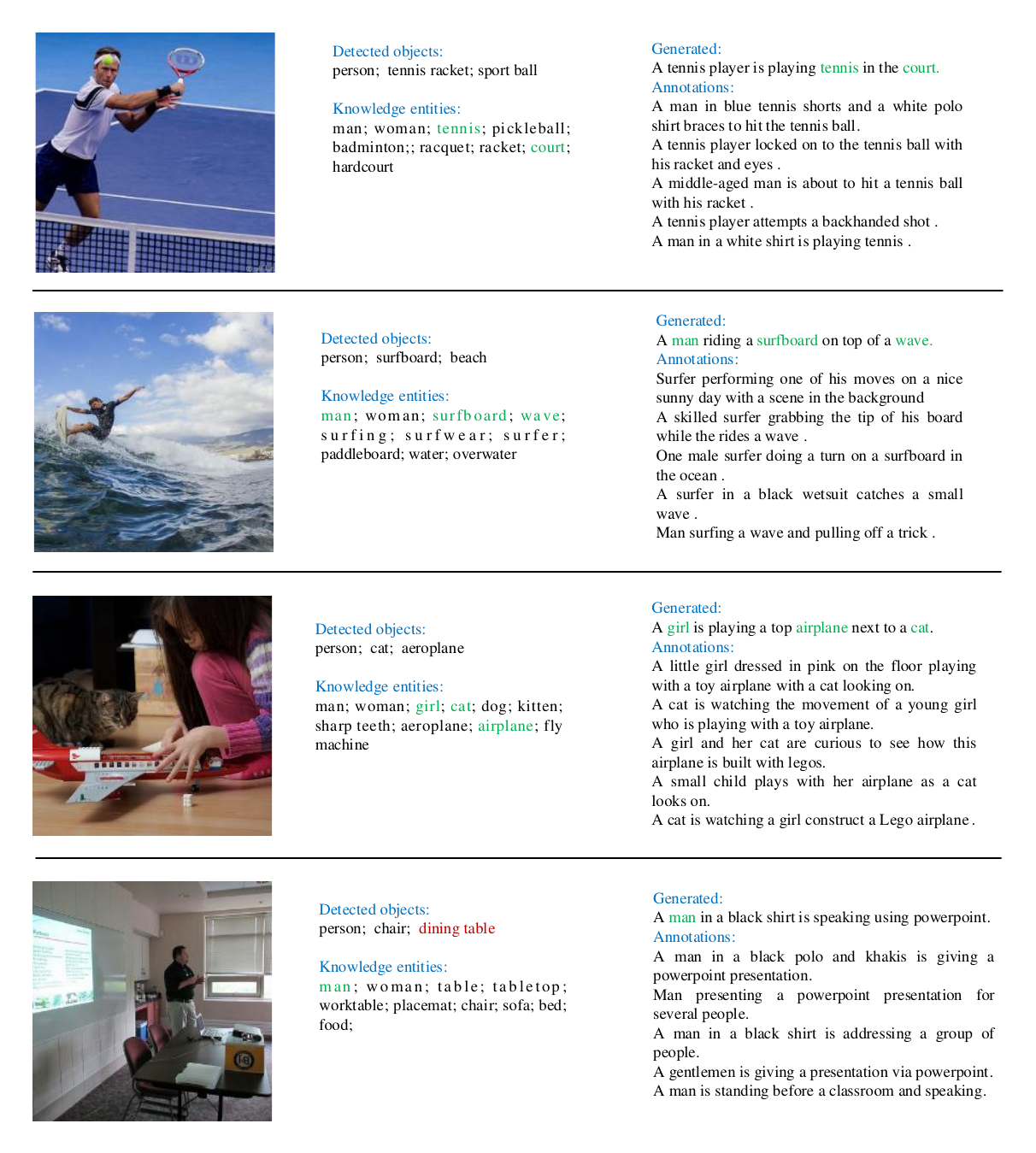}
  \caption{Some examples of captions generated by our full model. The 
  words with green color are the important words to be described.}
  \label{fig_sim}
\end{figure*}

\subsubsection{Qualitative result analysis}
\label{qra}
Furthermore, to evaluate the quality of captions generated by the entire
model and further verify the effectiveness of our proposed two scenarios,
we provide a qualitative analysis of the results. The visualization results
are shown in Fig.~\ref{fig_sim}. It can be seen that by combining word attention and
knowledge graph, our full model can generate more fine-grained captions,
as well as reveal more implicit aspects of the image,
which are not easily to be discovered by machines, but it seems not
difficult for humans. For example, look at the first picture,
our model can predict the object "court" even if it does not straightforwardly appear in the
ground-truth captions. In addition, the model prefers to use detected objects to describe
the image, of course these objects also appear in our constructed semantic knowledge
corpus.
However, like most existing models, limited by the caption length,
the proposed model does not perform well for complex images with multiple
objects. See the last picture, the model can't predict
the object "table". It is notable that the multi-object captioning involves another
more challenging study of artificial intelligence, i.e., dense captioning,
here we only consider generating a caption of a simple image. In general,
our model is suitable for the description of most scenes and achieves
performance compared to the state-of-the-art models.

\section{Coclusion}
\label{con}
In this paper, we explore to incorporate more useful semantic knowledge, including the
internal annotation knowledge and the external knowledge extracted from knowledge graph,
to reason about more accurate and meaningful captions.
We first propose a new text-dependent attention mechanism,
which we call word attention, to improve the correctness of basic visual
attention when generating sequential descriptions word-by-word. The special
word attention provides important semantic information
to the calculation of visual attention. We demonstrate that our proposed
model incorporating word attention as well as visual attention can significantly
improve the agreement between regions in image and words in sentence.
Then, in order to facilitate meaningful captioning and overcome the
problem of misprediction, we introduce a new strategy to incorporate 
commonsense knowledge extracted
form knowledge graph into the encoder-decoder framework. This has indeed
enhanced the generalization of our captioning model. Furthermore, we exploit
reinforcement learning to optimize our training process, thereby making a
significant improvement of the captioning performance. By combining the 
above mentioned several strategies, we achieve state-of-the-art performance on several standard
evaluation metrics.

In the future work, we expect to independently construct a more compact knowledge graph by using sentence
level and image level semantic information of a given instance to boost image captioning.

\section*{Acknowledgements}
This work is supported by the National Natural Science Foundation of
China (Nos. 61966004, 61663004, 61866004, 61762078), the Guangxi Natural
Science Foundation (Nos. 2019GXNSFDA245018, 2018GXNSFDA281009, 2017GXNSFAA198365),
the Guangxi “Bagui Scholar” Teams for Innovation and Research Project, the
Guangxi Talent Highland Project of Big Data Intelligence and Application,
Guangxi Collaborative Innovation Center of Multi-source Information
Integration and Intelligent Processing.

\bibliographystyle{splncs04}
\bibliography{ref}
\end{document}